\documentclass{llncs}
\usepackage{makeidx}  

\usepackage[tight]{subfigure}
\usepackage{wrapfig}
\usepackage{colortbl}
\usepackage{epsfig,latexsym}
\usepackage{url}
\usepackage{paralist}
\usepackage{amsmath,amssymb}
\usepackage[svgnames]{xcolor}
\usepackage{multirow}
\usepackage{booktabs}
\usepackage{dsfont}
\usepackage{xspace}
\usepackage{graphicx}
\usepackage{rotating}
\usepackage{times}
\usepackage{boxedminipage}

\usepackage{pifont}
\usepackage{booktabs}
\usepackage{xspace}
\usepackage{multirow}

\definecolor{Gray}{gray}{0.3}

\newcommand{\WS}{\ensuremath{\mathsf{WS}}\xspace}
\newcommand{\Tcheb}{\ensuremath{\mathsf{T}}\xspace}
\newcommand{\aTcheb}{\ensuremath{\mathsf{S}_{\textrm{aug}}}\xspace}

\newcommand{\Sgen}{\ensuremath{\mathsf{S}_{\textrm{gen}}}\xspace}
\newcommand{\thetasub}[1]{\theta_{\text{{#1}}}}
\newcommand{\SCALzero}{\ensuremath{\mathsf{S}_{\text{norm}}}\xspace}
\newcommand{\SCALtwo}{\aTcheb}

\newcommand{\SCALzerostar}{\ensuremath{\mathsf{S}^{\epsilon^\star}_{\text{norm}}}\xspace}
\newcommand{\SCALtwostar}{\ensuremath{\mathsf{S}^{\epsilon^\star}_{\textrm{aug}}}\xspace}

\newcommand{\rmnk}[0]{$\rho$MNK-landscape}

\renewcommand{\epsilon}{\varepsilon}

\newcommand{\ie}{i.e.}

\newcommand{\set}[1]{\left\{{#1}\right\}}

\newlength\ttt \newlength\tttt

\usepackage[normalem]{ulem}

\newcommand{\hide}[1]{}
\newcommand{\hideppsn}[1]{}

\makeatletter
\newcounter{roem}
\newcounter{caproem}
\renewcommand{\theroem}{\roman{roem}}
\renewcommand{\thecaproem}{\Roman{caproem}}
\newcommand {\c@org@eq }{}
\let \c@org@eq \c@equation
\newcommand {\org@theeq }{}
\let \org@theeq\theequation
\newcommand {\setroem }{
\let \c@equation \c@roem
\let \theequation \theroem }	
\newcommand {\setcaproem }{
\let \c@equation \c@caproem
\let \theequation \thecaproem }
\newcommand {\setarab }{
\let \c@equation \c@org@eq
\let \theequation \org@theeq }
\makeatother

\newcount\minute \newcount\hour \newcount\hourMins
\def\now{\minute=\time \hour=\time \divide \hour by 60 \hourMins=\hour \multiply\hourMins by 60
  \advance\minute by -\hourMins \zeroPadTwo{\the\hour}:\zeroPadTwo{\the\minute}}

\def\today{\the\year-\zeroPadTwo{\the\month}-\zeroPadTwo{\the\day}}
\def\zeroPadTwo#1{\ifnum #1<10 0\fi #1}

\begin{document}
\pagestyle{headings}  
%
\title{On the Impact of Multiobjective Scalarizing Functions}

\titlerunning{On the Impact of Multiobjective Scalarizing Functions}

\author{{Bilel Derbel}\inst{1,2} \hspace{0.5cm}
	{Dimo Brockhoff}\inst{1} \hspace{0.5cm}
	{Arnaud Liefooghe}\inst{1,2} \hspace{0.5cm}
	{S\'ebastien Verel}\inst{3} }

\authorrunning{Derbel, Brockhoff, Liefooghe, Verel}   
\institute{
		Inria Lille - Nord Europe, DOLPHIN project-team, France\\
\and
	Universit\'e Lille 1, LIFL, UMR CNRS 8022, France\\
\and
Universit\'e du Littoral C\^ote d'Opale, LISIC, France
}

\maketitle 

\vspace{-3ex}
\begin{abstract}
Recently, there has been a renewed interest in decomposition-based approaches for evolutionary multiobjective optimization. However, the impact of the choice of the underlying scalarizing function(s) is still far from being well understood. In this paper, we investigate the behavior of different scalarizing functions and their parameters. We thereby abstract firstly from any specific algorithm and only consider the difficulty of the single scalarized problems in terms of the search ability of a $(1+\lambda)$-EA on biobjective NK-landscapes. Secondly, combining the outcomes of independent single-objective runs allows for more general statements on set-based performance measures. Finally, we investigate the correlation between the opening angle of the scalarizing function's underlying contour lines and the position of the final solution in the objective space. Our analysis is of fundamental nature and sheds more light on the key characteristics of multiobjective scalarizing functions.
\end{abstract}

\vspace{-4.5ex}
\section{Introduction}\label{sec:intro}
Multiobjective optimization problems occur frequently in practice and evolutionary multiobjective optimization (EMO) algorithms have been shown to be well-applicable for them---especially if the problem under study is nonlinear and/or derivatives of the objective functions are not available or meaningless. Besides the broad class of Pareto-dominance based algorithms such as NSGA-II or SPEA2, a recent interest in the so-called \emph{decomposition-based algorithms} can be observed. Those decompose the multiobjective problem into a set of single-objective, `scalarized' optimization problems. 
Examples of such algorithms include MSOPS \cite{hugh2003b}, MOEA/D \cite{zl2007a}, and their many variants. We refer to \cite{gpf2013a} for a recent overview on the topic. The main idea behind those algorithms is to define a set of (desired) search directions in objective space and to specify the scalarizing functions corresponding to these directions. The scalarizing functions can then be solved independently (such as in the case of MSOPS), or in a dependent manner (like in MOEA/D where the recombination and selection operators are allowed to use information from the solutions maintained in neighboring search directions).

Many different scalarizing functions have been proposed in the literature, see e.g.~\cite{miet1999a} for an overview. Well-known examples are the weighted sum and the (augmented) weighted Chebychev functions, where the latter has an inherent parameter that controls the shape of the lines of equal function values in objective space.
Especially with respect to decomposition-based EMO algorithms, it has been reported that the choice of the scalarizing function and their parameters has an impact on 
the search process~\cite{gpf2013a}. 
Moreover, it has been noted that adapting the scalarizing function's parameters during the search can allow improvement 
over having a constant set of scalarizing functions~\cite{istn2009a}. Although several studies on the impact of the scalarizing function 
have been conducted in recent years, e.g.~\cite{IshibuchiAN13}, to the best of our knowledge, all of them investigate it 
on a concrete EMO algorithm and on the quality of the resulting solution sets when more than one scalarizing function is optimized (typically as mentioned above, in a dependent manner). Thereby, the focus is not in understanding why those performance differences occur but rather in observing them and trying to improve the global algorithm. However, we believe that it is more important to first understand thoroughly the impact of the choice of the scalarizing function for a \emph{single} search direction before analyzing more complicated algorithms such as MOEA/D-like approaches with specific neighboring structures, recombination, and selection operators.
In this paper, we fundamentally investigate the impact of the choice of the scalarizing functions and their parameters on 
the search performance, independently of any known EMO algorithm. Instead, we consider one of the most simple 
single-objective scalarizing search algorithms, \ie, a $(1+\lambda)$-EA with standard bit mutation, as an example of a local search algorithm that optimizes a single scalarizing function, corresponding to a single search direction in the objective space.
Experiments are conducted on well-understood bi-objective \rmnk s.

More concretely, we look experimentally at the impact of the parameters of a generalized scalarizing function (which covers the special cases of the weighted sum and augmented Chebychev scalarizing functions) in terms of the position (angle/direction) reached by the final points, as well as their quality with respect to the Chebychev function. We then consider how the opening of the cones that describe the lines of equal scalarizing function values can provide a theoretical explanation for the impact of the final position of the obtained solutions in objective space. We also investigate the resulting \emph{set quality} in terms of hypervolume and $\epsilon$-indicator if several scalarizing $(1+\lambda)$-EAs are run independently for different search directions in the objective space.
Finally, we conclude our findings with a comprehensive discussion of promising research lines.

\vspace{-2ex}
\section{Scalarizing Functions}
\label{sec:scalaringFunction}

We consider the maximization of two objectives $f_1$, $f_2$ that map search points $x\in X$ to an objective vector $f(x) = (f_1(x), f_2(x)) = (z_1,z_2)$ in the so-called objective space $f(X)$. A solution $x$ is called {\em dominated} by another solution $y$ if $f_1(y) \geq f_1(x)$, $f_2(y) \geq f_2(x)$, and for at least one $i$, $f_i(y)>f_i(x)$ holds. The set of all solutions, not dominated by any other, is called Pareto set and its image Pareto front.

Many ways of \emph{decomposing} a multiobjective optimization problem into a (set of) single-objective \emph{scalarizing functions} exist, including the prominent examples of \emph{weigh\-ted sum} (\WS), {\em weighted Chebychev} (\Tcheb), or {\em augmented weighted Chebychev}~(\aTcheb)~\cite{miet1999a}. For most of them, theoretical results, especially about which Pareto-optimal solutions are attainable, exist \cite{miet1999a,kali2001} but they are typically of too general nature to allow for statements on the actual search performance of (stochastic) optimization algorithms. Instead, we are here not interested in any particular scalarizing function, but rather in understanding which general properties of them influence the search behavior of EMO algorithms. 
We argue by means of experimental investigations that it is not the actual choice of the scalarizing function or their parameters that makes the difference in terms of performance, 
but rather the general properties of the resulting lines of equal function values.
To this end, we consider the minimization of 
the following general scalarizing function that covers the special cases of \WS\footnote{Contrary to the standard literature, our formalization assumes minimization and we therefore have included the utopian point $\bar{z}$ that is typically assumed to be $\bar{z} = (0,0)$ for minimization.}, \Tcheb, and \aTcheb functions:
\begin{equation*}\label{eq:scal}
\begin{aligned}
	\Sgen(z) &= \alpha \cdot \max \left\{ \lambda_1 \cdot |\bar{z}_1 - z_1|, \lambda_2 \cdot |\bar{z}_2 - z_2| \right\} + \epsilon \left( w_1 \cdot |\bar{z}_1 - z_1| + w_2 \cdot |\bar{z}_2 - z_2| \right)
\end{aligned}
\end{equation*}
where $z=(z_1,z_2)$ is the objective vector of a feasible solution, $\bar{z} = (\bar{z}_1, \bar{z}_2)$ a utopian point, $\lambda_1, \lambda_2, w_1,$ and $w_2 > 0$ scalar weighting coefficients indicating a search direction in objective space, and $\alpha\geq 0$ and $\epsilon\geq 0$ parameters to be fixed. For more details about the mentioned scalarizing functions and their relationship, we refer to Table~\ref{tab:scalarizingfunctions}.

In the following, we also consider a case of $\Sgen$ that combines \WS and \Tcheb with a single parameter $\epsilon$: the normalized $\SCALzero(z) = (1-\epsilon) \Tcheb(z) + \epsilon \WS(z)$ where $\alpha=1-\epsilon$ and $\epsilon\in [0,1]$.
For optimizing in a given search direction $(d_1,d_2)$ in objective space, we follow~\cite{hugh2003b,bwt2012a} and set $\lambda_i = 1/d_i$.\footnote{The pathologic cases of directions parallel to the coordinates are left out to increase readability.} In addition, we refer to the direction angle as $\delta=\arctan(d_1/d_2)$. For the case of \SCALzero, we furthermore choose $w_1=\cos(\delta)$ and $w_2=\sin(\delta)$ (thus, $w_1^2 + w_2^2 = 1$) for the weighted sum part in order to normalize the search directions in objective space uniformly w.r.t. their \emph{angles}.
Though, in many textbooks you can find statements like ``$\epsilon$ has to be chosen small (enough)'', we do not make such an assumption but want to understand which influence  $\epsilon$ has on the finally obtained solutions and how it introduces a trade-off between the Chebychev approach and a weighted sum. For the question of how small $\epsilon$ should be chosen to find all Pareto-optimal solutions in exact biobjective discrete optimization, we refer to \cite{dgk2011a}.

\begin{table*}[t]
	\centering
	\caption{\label{tab:scalarizingfunctions} Overview of the considered scalarizing functions, and the corresponding angles of the lines of equal function values with the standard Pareto dominance cone.\vspace*{-0.7em}}
\begin{scriptsize}
	\begin{tabular}{lp{1.8cm}p{3.8cm}l}
	\toprule
	scalar function & parameters in \Sgen           & opening angles        & reference\\
	                 \midrule
	 $\WS(z) = w_1 |\bar{z}_1 - z_1|  + w_2 |\bar{z}_2 - z_2|$ & $\alpha=0$, $\epsilon=1$ & 
																	\begin{minipage}[t]{4.2cm}
																		$\thetasub{1} = \arctan\big(-\frac{w_1}{w_2}\big)$\\
																		$\thetasub{2} = \frac{\pi}{2} + \arctan\big(\frac{w_1}{w_2}\big)$
																	\end{minipage} & \cite[Eq.~3.1.1]{miet1999a}\\[0.5em]
	$\Tcheb(z) = \max \{ \lambda_1 |\bar{z}_1 - z_1|, \lambda_2 |\bar{z}_2 - z_2| \}$ & $\alpha=1$, $\epsilon=0$ & 
																		\begin{minipage}[t]{4.2cm}
																			$\thetasub{1} = 0$\\
																		$\thetasub{2} = \pi/2$
																		\end{minipage} & \cite[Eq.~3.4.2]{miet1999a}\\[1.5em]
	
		$\aTcheb(z) = \Tcheb(z) + \epsilon \left( |\bar{z}_1 - z_1| + |\bar{z}_2 - z_2| \right)$
	& $\alpha=1$,\;\;\;\;\;\;\;\;\;\;\;\; $w_1=w_2=1$ & 
																		\begin{minipage}[t]{4.2cm}
																			$\thetasub{1} = \arctan\big(-\frac{\epsilon}{\lambda_1 + \epsilon}\big)$\\
																			$\thetasub{2} = \frac{\pi}{2} + \arctan\big(\frac{\epsilon}{\lambda_2 + \epsilon}\big)$
																		\end{minipage} & \cite[Eq.~3.4.5]{miet1999a}\\[2.5em]
 $\SCALzero(z) = (1-\epsilon) \Tcheb(z) + \epsilon \WS(z)$ & $\alpha=1-\epsilon$,\; $w_i=1/\lambda_i$ & 
																		\begin{minipage}[t]{4.2cm}
	           													$\thetasub{1} = \arctan(-\frac{\epsilon w_1}{(1-\epsilon)\lambda_2 + \epsilon w_2})$\\
	           													$\thetasub{2} = \frac{\pi}{2} + \arctan(\frac{\epsilon w_2}{(1-\epsilon)\lambda_1 + \epsilon w_1})$
	           												\end{minipage} & here\\[1.5em]
	\bottomrule
	\end{tabular}
	\end{scriptsize}
	\vspace*{-1.8em}
\end{table*}

As mentioned above, one important property of a scalarizing function turns out to be the shape of its sets of equal function values, which are known for the \WS, \Tcheb, and \aTcheb 
functions \cite{miet1999a}.
However, no description of the equi-function-value lines for the general scalarizing function \Sgen 
has been given so far. We think that it is necessary to state those opening angles explicitly in order to gain a deeper intuitive understanding of the above scalarizing approaches and related concepts such as the R2 indicator~\cite{bwt2012a} or more complicated scalarizing algorithms such as MOEA/D~\cite{zl2007a}. Moreover, it allows us to investigate how a linear combination of weighted sum and Chebychev functions
affect the search behavior of decomposition-based algorithms. 
The following proposition, proven in the accompanying report \cite{supplMatos14}, states these opening angles $\theta_i$ between the equi-utility lines and the $f_1$-axis, see also Fig.~\ref{fig:dynamic} for some examples. 
\begin{proposition}\label{thm:angles}
	Let $\bar{z}$ 
	be a utopian point, $\lambda_1, \lambda_2, w_1,$ and $w_2 > 0$ scalar weighting coefficients, $\alpha\geq 0$ and $\epsilon\geq 0$, where at least one of the latter two is positive. Then, the polar angles between the equi-utility lines 
	of~~\Sgen 
	and the $f_1$-axis are $\thetasub{1} = \arctan(-\frac{\epsilon w_1}{ \alpha\lambda_2 + \epsilon w_2})$ and $\thetasub{2} = \frac{\pi}{2} + \arctan(\frac{\epsilon w_2}{\alpha\lambda_1 + \epsilon w_1})$.
\end{proposition}

\section{Experimental Design}
\label{sec:exp}

This section presents the experimental setting allowing us to analyze the scalarizing approaches introduced above on bi-objective \rmnk s.
The family of \rmnk s constitutes a problem-independent model used for constructing multiobjective multimodal landscapes with objective correlation~\cite{verel2012}.
A bi-objective \rmnk\ aims at maximizing an objective function vector $f : \set{0,1}^n \rightarrow [0,1]^2$. A correlation parameter~$\rho$ defines the degree of conflict between the objectives. We investigate a random instance for each parameter combination given in Table~\ref{tab:param}.

\begin{table}[t]
\caption{Parameter setting.}
\vspace*{-5ex}
\begin{scriptsize}
\begin{center}
\begin{tabular}{rl|c|c}
\hline
\multicolumn{2}{c}{scalarizing functions}& \multicolumn{1}{c}{\rmnk s}&$(1+\lambda)$-EA\\
\hline
$\bar{z}$	&	$=~(1,1)$ & $\rho \in \set{-0.9,-0.8,\ldots,0.0,\ldots,0.9}$ &$\lambda=n$\\
$\delta$	&	$=~j \cdot 10^{-2} \cdot \frac{\pi}{2}$, $~~j \in [\![ 1,99 ]\!]$& $m=2$ &bit-flip rate $=1/n$\\
\SCALzero: & $\epsilon~= ~\ell\cdot 10^{-2}$; $~~\ell \in [\![ 0,100]\!]$ & $n=128$ & stopped after\\
\SCALtwo:	&	$\epsilon ~=~ \ell \cdot 10^{-k};~~\ell \in [\![ 0,10]\!]; k\in [\![-1,2]\!]$ & $k=4$ & $n$ iterations\\ \hline

\end{tabular}
\end{center}
\label{tab:param}
\end{scriptsize}
\vspace*{-2em}
\end{table}

We investigate the two scalarizing functions \SCALzero and \SCALtwo of Table~\ref{tab:scalarizingfunctions} with different parameter settings
for the weighting coefficient vector and the $\epsilon$ parameter, as reported in Table~\ref{tab:param}. In particular, the \WS (resp. \Tcheb) function corresponds to \SCALzero with $\epsilon=1$ (resp. $\epsilon=0$).
The set of weighting coefficient \emph{direction angles} $\delta_j$ with respect to the f$_1$-axis ($j \in \set{1, \dots, 99}$) are uniformly defined with equal distances in the angle space. For both functions, we set $\lambda_1=1/\cos(\delta_j)$, and $\lambda_2=1/\sin(\delta_j)$. We recall that for \SCALzero, $w_i=1/\lambda_i$, and for \SCALtwo, $w_i = 1$. 
To evaluate the relative and the joint performance of the considered scalarizing functions,
we investigate the dynamics and the performance of a randomized local search, a simple $(1+\lambda)$-EA. 
After initially drawing a random solution, at each iteration, $\lambda$ offspring solutions are generated by means of an independent bit-flip mutation, where each bit of the parent solution is independently flipped with a rate $1/n$.
The solution with the best (minimum) scalarizing function value among parent and offspring is chosen for the next iteration. For each configuration, $30$ independent executions are performed.
Due to space limitations, we shall only show a representative subset of settings allowing us to state our findings. More exhaustive results can be found in~\cite{supplMatos14}.

\begin{figure*}[t]
\includegraphics[width=0.33\columnwidth]{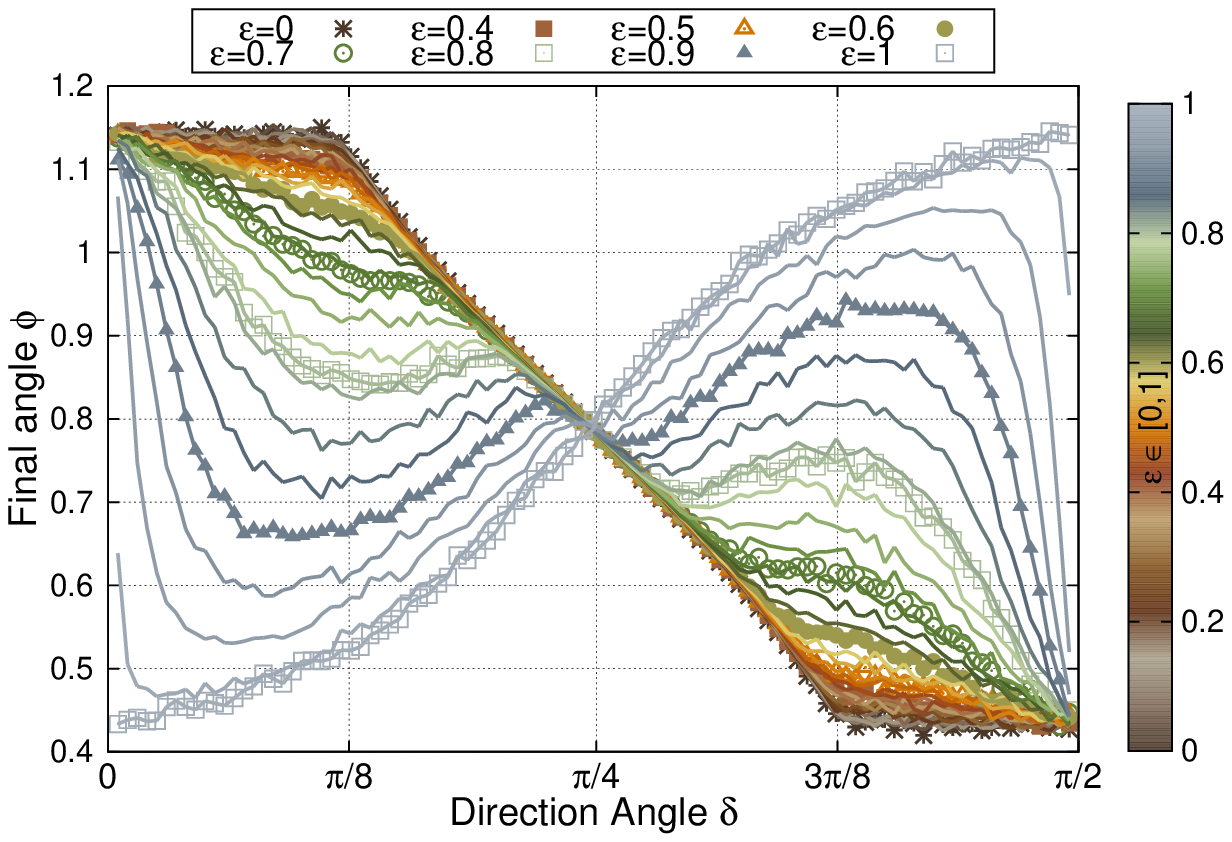}
\includegraphics[width=0.33\columnwidth]{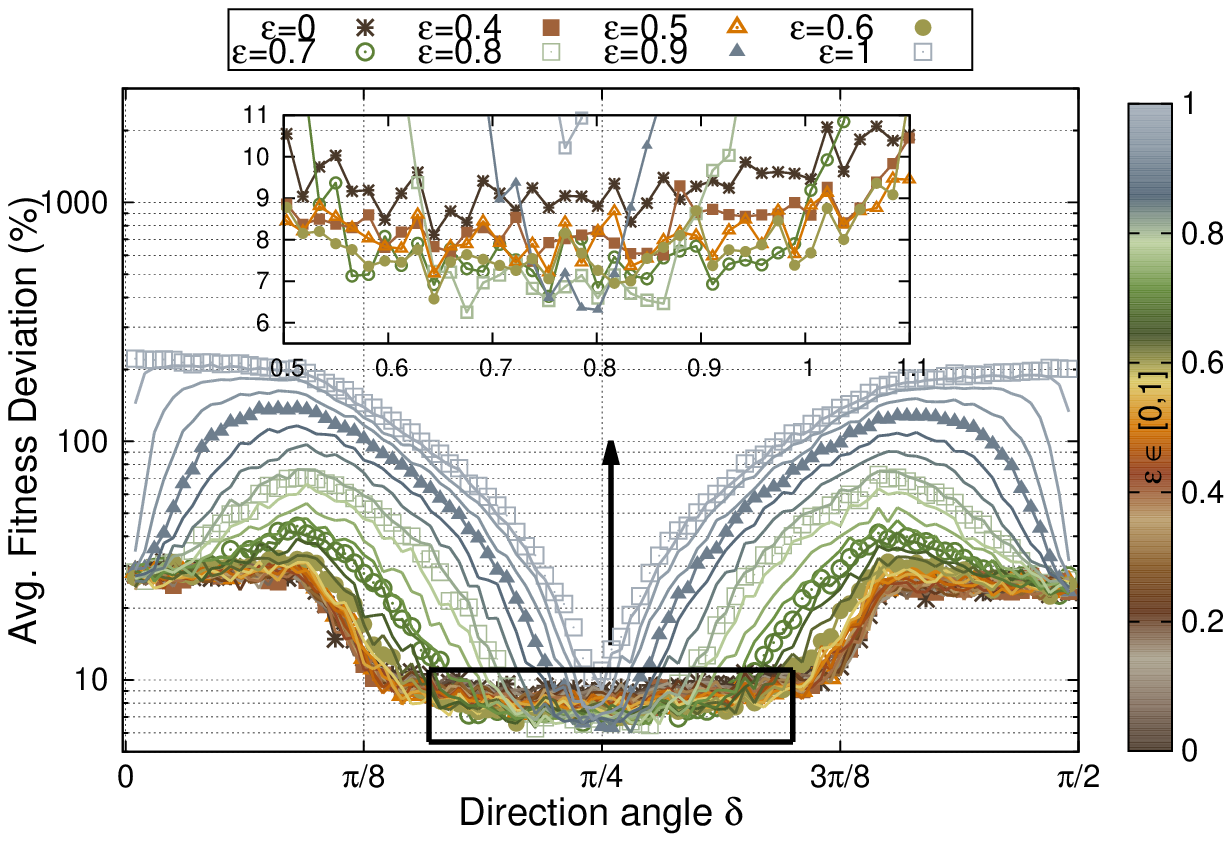}
\includegraphics[width=0.33\columnwidth]{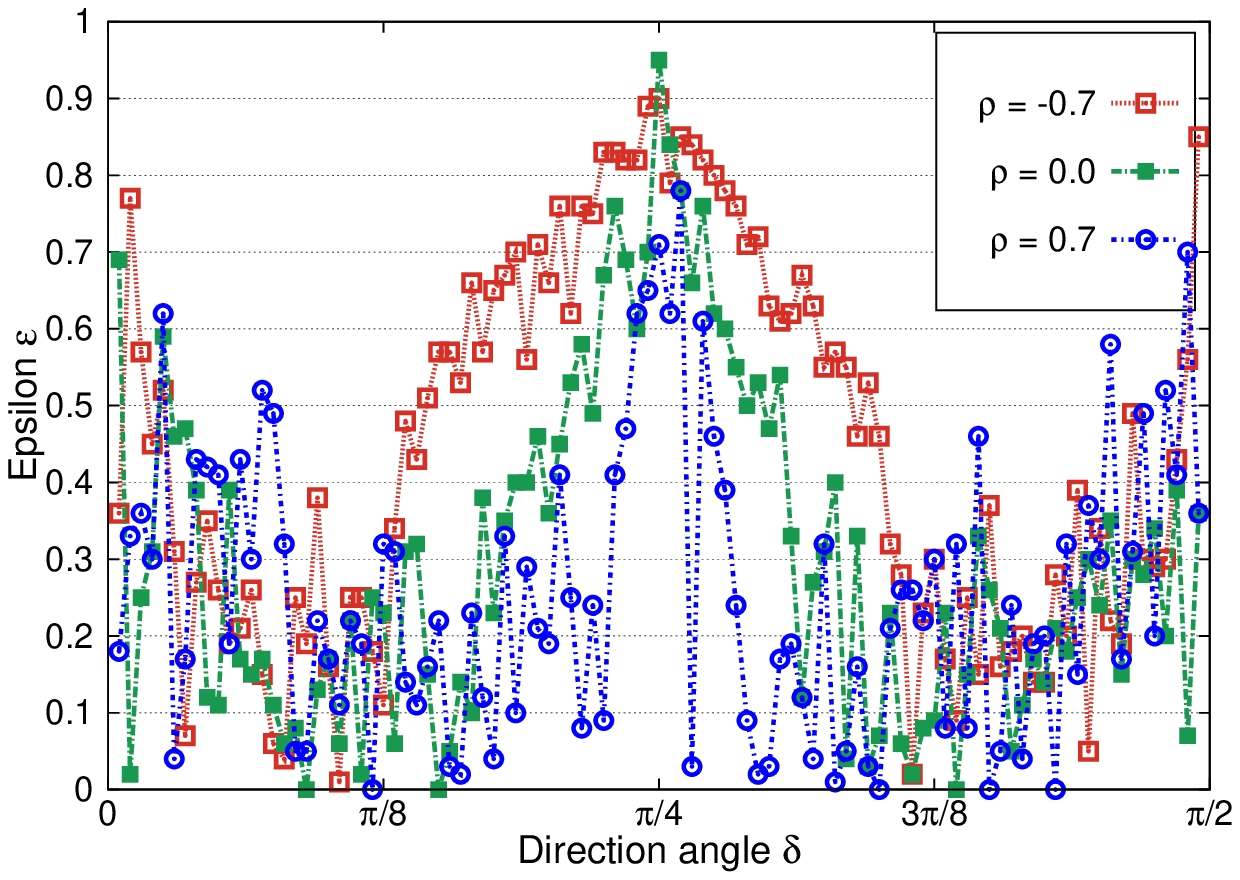}
\vspace{-5ex}
\caption{Left: Angles of final solutions for $\SCALzero$ and $\rho=-0.7$ as a function of $\delta$. Middle: Average deviation to best as a function of weight vector for $\SCALzero$ and $\rho=-0.7$. Right: $\epsilon$-values providing the smallest deviation for every fixed direction in $\SCALzero$ and  $\rho \in \set{-0.7,0.0,0.7}$.\vspace{-3ex}}
\label{fig:angledeviation}
\end{figure*}

\section{Single Search Behavior}
\label{sec:conc}

This section is devoted to the study of the optimization paths followed by \emph{single} independent $(1+\lambda)$-EA runs for each direction angle $\delta$ and parameter $\epsilon$ of a scalarized problem. 
In particular, we study the final solution sets reached by the $(1+\lambda)$-EA in terms of diversity and convergence and give a sound explanation on how the search behaviour is related to the lines of equal function values of the scalarizing functions.

\subsection{Diversity: Final Angle}

In Fig.~\ref{fig:angledeviation} (Left), we examine the average \emph{angle} of the \emph{final} solution reached by the algorithm with respect to the $f_1$-axis using $\SCALzero$. The final angle of solution $x$ is defined as $\phi(x)=\arctan(f_2(x)/f_1(x))$. It informs about the actual 
\emph{direction} followed by the search process. 
We can see that the final solutions are in symmetric positions with respect to direction angle~$\pi/4$. This is coherent with the symmetric nature of \rmnk s~\cite{verel2012}. For $\WS$ ($\epsilon=1$), every single direction angle infers a different final angle. 
For $\Tcheb$ ($\epsilon=0$), the extreme direction angles end up reaching `similar' regions of the objective space. These regions actually correspond to the lexicographically optimal points of the Pareto front, which is because of the choice of the utopian point that lies beyond them. Without surprise, we can also see that \Tcheb and \WS do not always allow to approach the same parts of the Pareto front when using the same direction angle. 

When varying $\epsilon$ for a fixed $\delta$, the search process is able to span a whole range of positions that are achieved by either \Tcheb or \WS but for variable $\delta$ values. Actually, when considering the direction angle being in the middle (i.e.\ $\delta\approx\pi/4$), the choice of $\epsilon$ does not substantially impact the search direction---because \Tcheb and \WS do allow to move to similar regions in this case. However, as the direction angle goes away from the middle, the influence of $\epsilon$ grows significantly; and the search direction is drifting in a whole range of values. This indicates that the choice of $\delta$ is not the only feature that determines the final angle but also the choice of $\epsilon$ highly matters: For some specific $\epsilon$-values, the direction angles allow to distribute final angles fairly between the two lexicographically optimal points of the Pareto front---in the sense that each direction angle is inferring a different final angle, just like what we observe for $\WS$. For some other $\epsilon$-values, however, it may happen that the final angles are similar for two different direction angles. In particular, this is the case for large $\epsilon$-values in $\SCALzero$, for which $\WS$ has more impact than~$\Tcheb$. We remark that equivalent conclusions can be drawn when examining $\SCALtwo$, which we do not detail here due to lack of space.

The distribution of final directions is tightly related to the diversity of solutions computed by different independent single (1+$\lambda$)-EAs. As it will be discussed later, this is of crucial importance from a multiobjective standpoint, since diversity in the objective space is crucial to approach different parts of the Pareto front.

\vspace{-1.5ex}
\subsection{Convergence: Relative Deviation to Best}\label{sec:bestdeviation}
In the following, we examine the impact of the scalarizing function parameters on the performance of the $(1+\lambda)$-EA in terms of convergence to the Pareto front. For that purpose, we compute, for every direction angle $\delta$, the best-found objective vector $z^\star_{\delta,\Tcheb}$ corresponding to the best (minimum) fitness value with respect to $\Tcheb$, over all experimented parameter combinations and over all simulations we investigated. For both functions $\SCALzero$ and $\SCALtwo$, we consider the final objective vector $z$ obtained for every direction angle $\delta$ and every $\epsilon$-value. We then compute the relative deviation of $z$ with respect to $z^\star_{\delta,\Tcheb}$, which we define as follows: $\Delta(z)=(\Tcheb(z)-\Tcheb(z^\star_{\delta,\Tcheb}))/\Tcheb(z^\star_{\delta,\Tcheb})$. 
Notice that this relative deviation factor is computed with respect to the $\Tcheb$ function, which is to be viewed as a reference measure of solution quality. This value actually informs about the performance of the $(1+\lambda)$-EA for a fixed direction angle, but variable $\epsilon$-values.

In Fig.~\ref{fig:angledeviation} (Middle), we show the average relative deviation to best as a function of direction angles ($\delta$) for different $\epsilon$-values. To understand the obtained results, one has to keep in mind the results discussed in the previous section concerning the final angles inferred by a given parameter setting. In particular, since $\WS$ and  $\Tcheb$ do not infer similar final angles, the final computed solutions lay in different regions of the objective space. Also, for the extreme direction angles, different ranges of $\epsilon$ imply different final angles. Thus, it is with no surprise that the average relative deviation to best can be substantial in such settings. However, the situation is different when considering direction angles in the middle ($\delta\approx \pi/4$). In fact, we observe that for such a configuration, the $\epsilon$-value does \emph{not} have a substantial effect on final angles, \ie, final solutions lie in similar regions of the objective space. Hence, one may expect that the search process has also the same performance in terms of average deviation to best. This is actually \emph{not} the case since we can observe that the value of $\epsilon$ has a significant impact on the relative deviation for the non-extreme direction angles. To better illustrate this observation, we show, in Fig.~\ref{fig:angledeviation} (Right), the $\epsilon$-value providing the minimum average relative deviation to best as a function of every direction angle. We clearly see that the best performances of the $(1+\lambda)$-EA for different direction angles are not obtained with the same $\epsilon$-value.

\subsection{Understanding the Impact of the Opening Angle}
\label{sec:predic}

In this section, we argue that the dynamics of the search process observed previously is rather independent of the scalarizing function under consideration or its parameters.
Instead, we show that the search process is guided by the positioning of the lines of equal function values in the objective space---described by the opening angle, \ie, the angle between the line of equal function values and the $f_1$-axis (cf.\ Proposition~\ref{thm:angles}).

\begin{figure*}[t]
\begin{center}
\includegraphics[width=0.33\columnwidth]{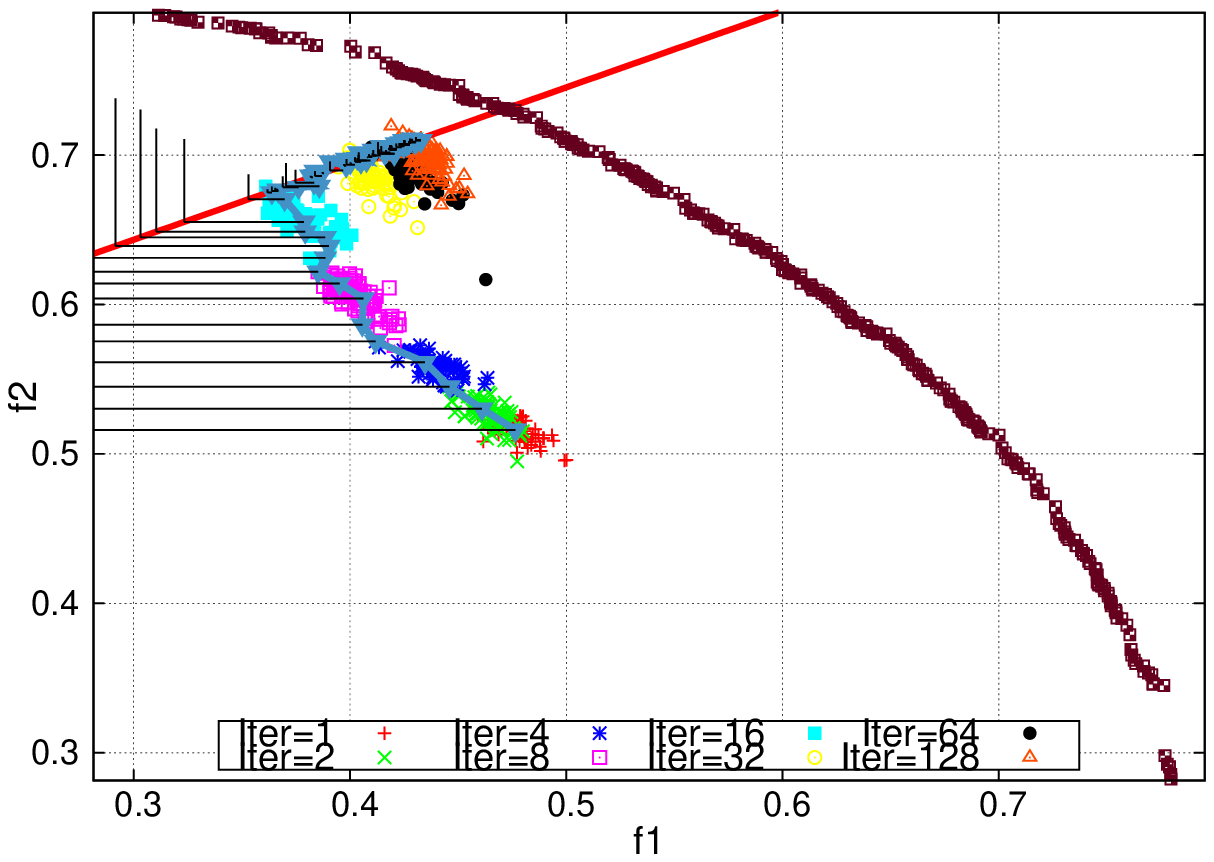}\hspace{-1ex}
\includegraphics[width=0.33\columnwidth]{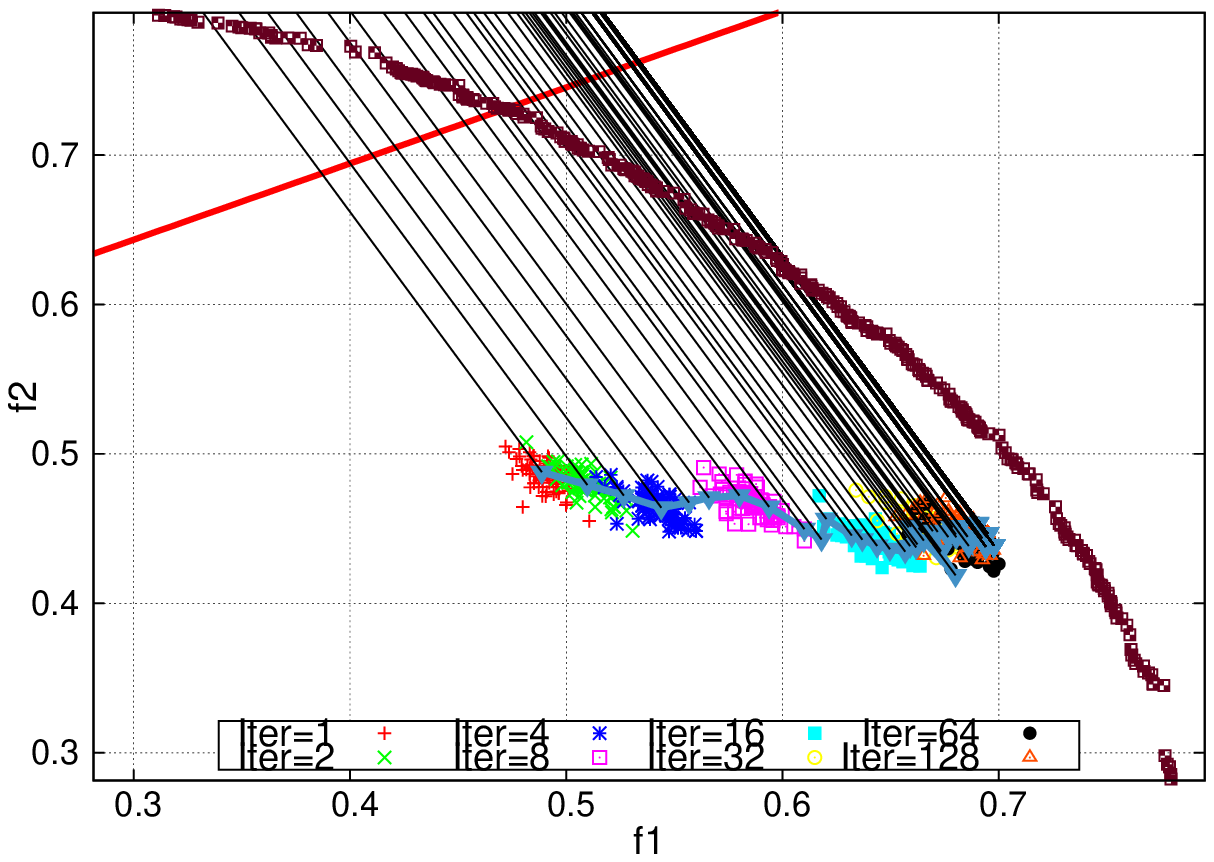}\hspace{-1ex}
\includegraphics[width=0.33\columnwidth]{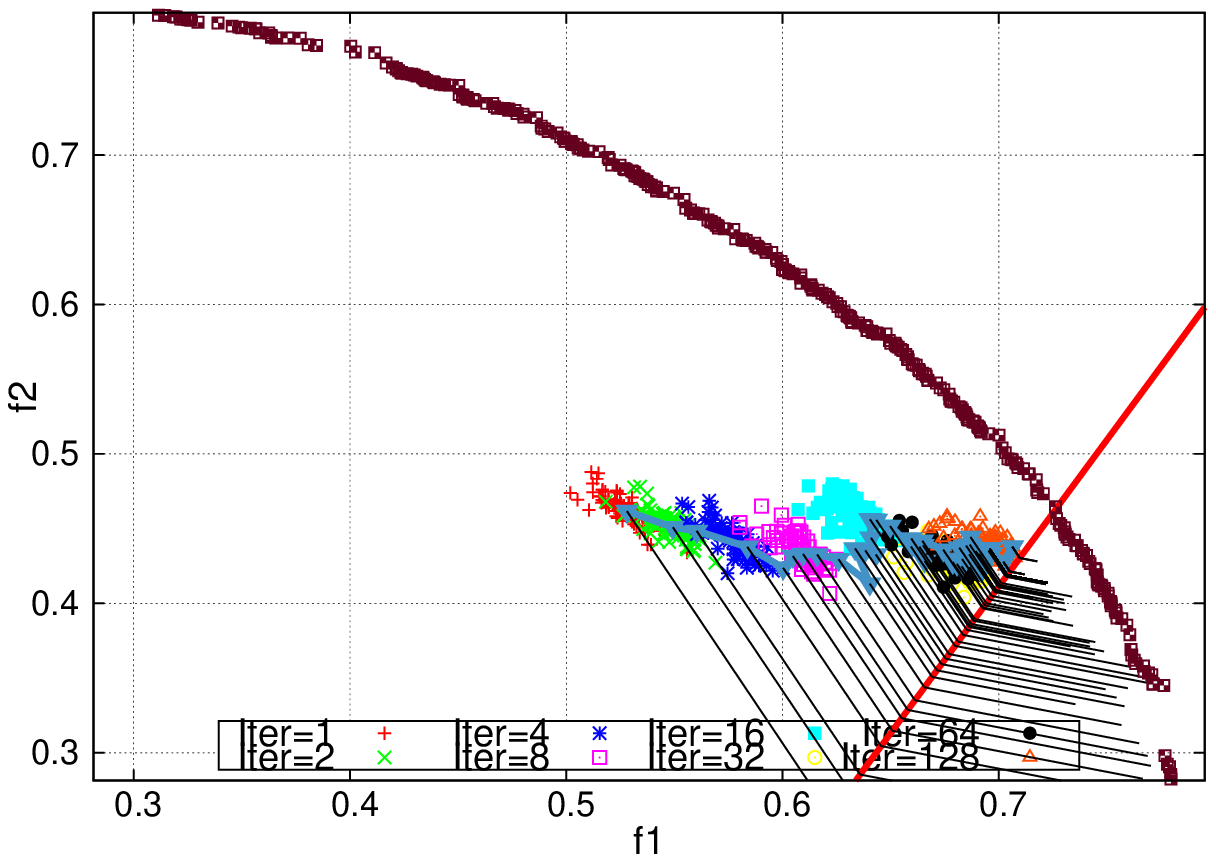}
\vspace{-3.5ex}
\caption{Exemplary runs of the $(1+\lambda)$-EA for different direction angles $\delta$ (straight line) and different $\epsilon$-values (\SCALzero, $\rho=-0.7$). Shown are the best known Pareto front approximation, the offspring at some selected generations, the evolution of the parent, and the lines of equal function values. Left: $\epsilon=0$, $\delta=\frac{3}{10}\cdot \frac{\pi}{2}$. Middle: $\epsilon=1$, $\delta = \frac{3}{10}\cdot\frac{\pi}{2}$. Right: $\epsilon=0.6$, $\delta=\frac{7}{10}\cdot \frac{\pi}{2}$.\vspace{-5ex}}
\label{fig:dynamic}
\end{center}
\end{figure*}

Fig.~\ref{fig:dynamic} shows three typical exemplary executions of the $(1+\lambda)$-EA in the objective space for different parameter settings. The typical initial solution maps around the point $z=(0.5, 0.5)$ in the objective space, which is the average objective vector for a random solution of \rmnk s.
The evolution of the current solution can be explained by the combination of two effects.
The first one is given by the independent bit-flip mutation operator, that produces more offspring in a particular direction compared to the other ones,
due to the underlying characteristics of the \rmnk~under consideration. The second one is given by the lines of equal function values, \ie, the current solution moves perpendicular to the iso-fitness lines, following the gradient direction in the objective space.
We can remark that the search process is mainly guided by the lower part of the cones of equal function values when the direction is above the initial solution,
and \textit{vice versa}.
When the direction angle $\delta$ is smaller (resp. larger) than $\pi / 4$, 
the dynamics of the search process are better captured by the opening angle~$\theta_1$ (resp.~$\theta_2$), defined between the equi-fitness lines and the $f_1$-axis.
Geometrically, the optimal solution with respect to a scalarizing function should correspond to the intersection of one of the `highest' lines of equal fitness values in the gradient direction and the feasible region of the objective space.
Although the above description is mainly intuitive, a more detailed analysis can support this general idea.

\begin{figure*}[t]
\begin{center}
\begin{tabular}{ccc}
\includegraphics[width=0.33\textwidth]{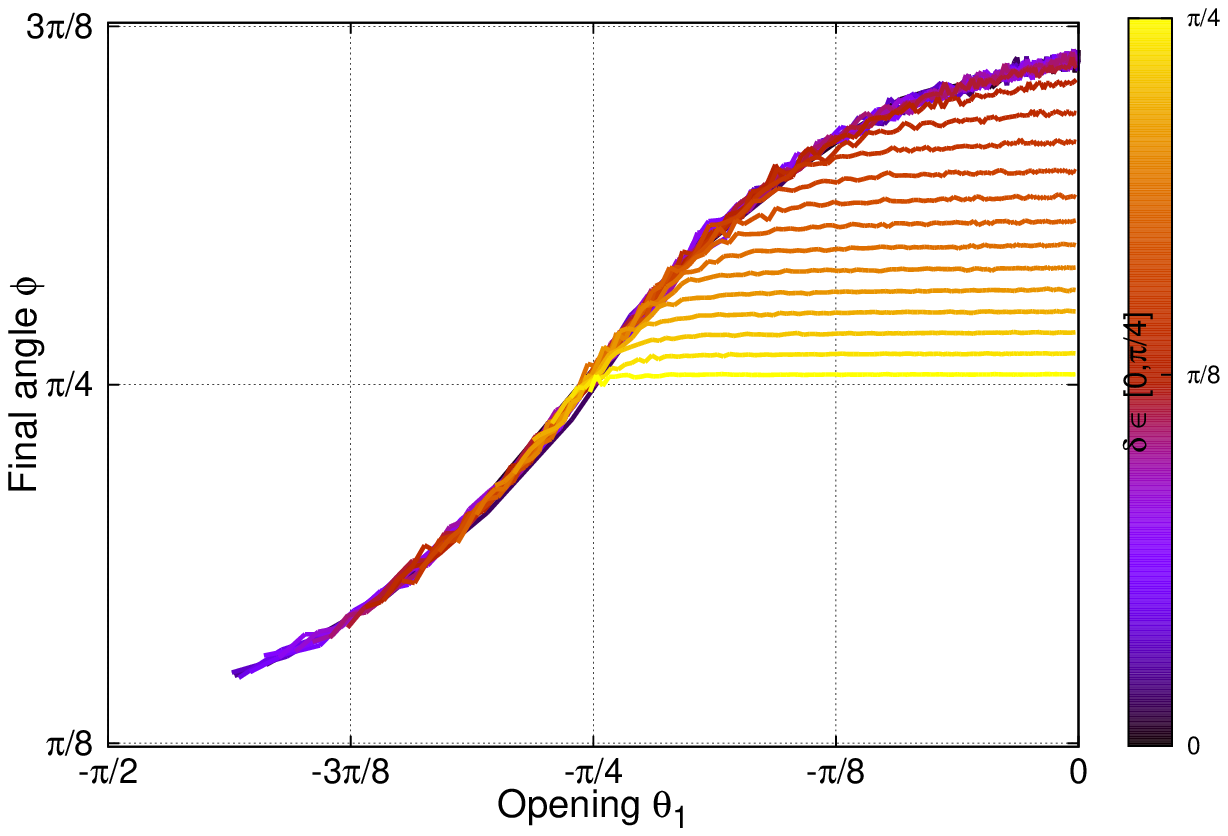} & 
\includegraphics[width=0.33\textwidth]{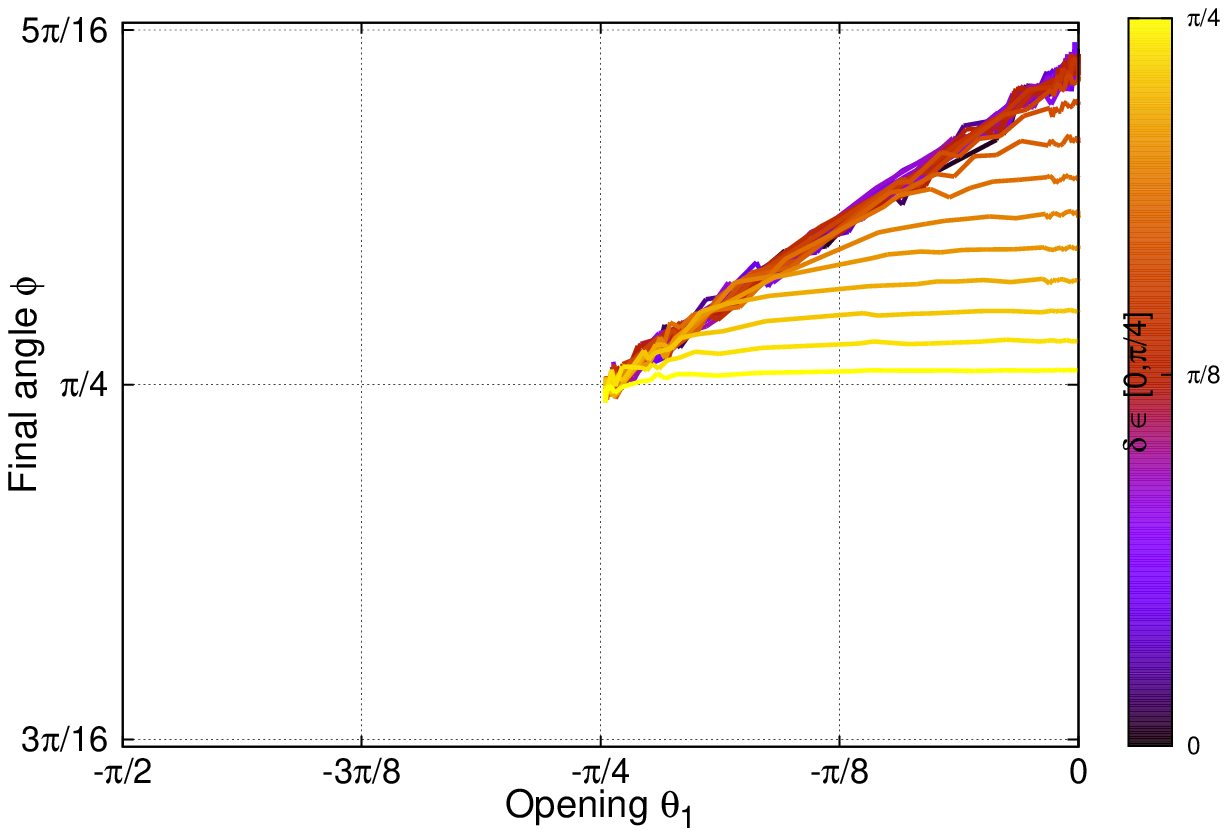} & 
\includegraphics[width=0.33\textwidth]{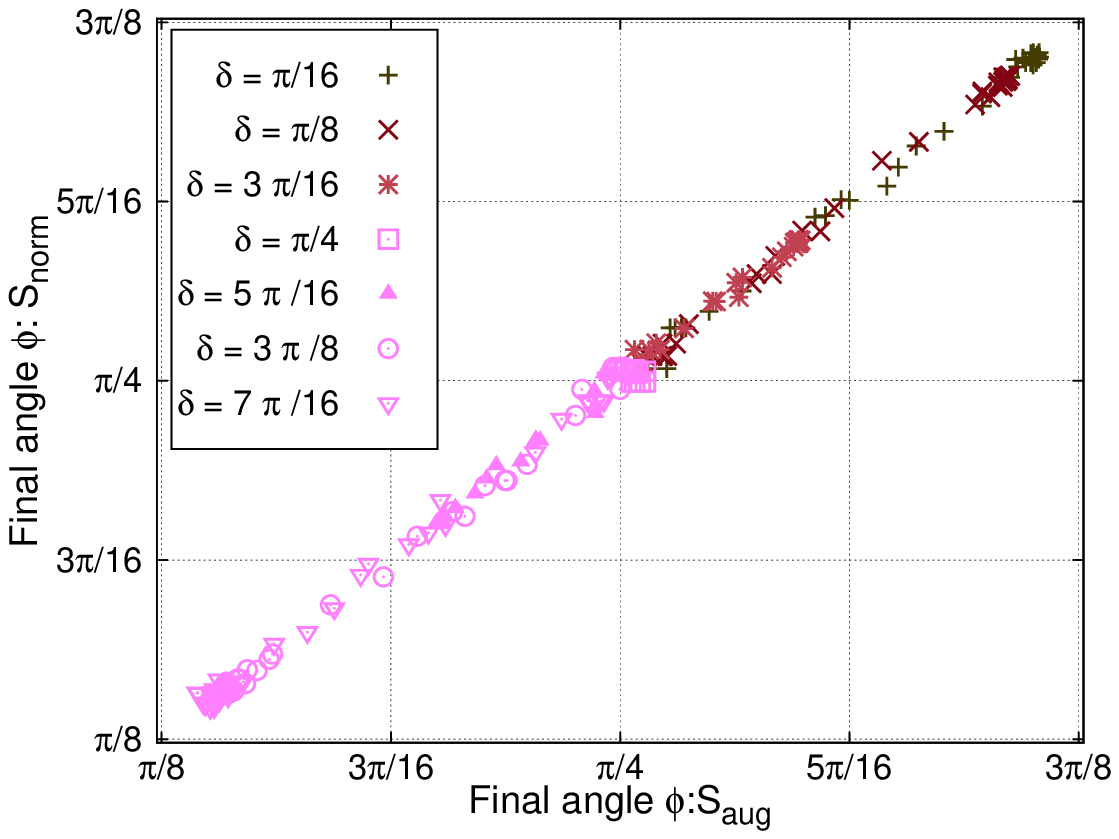} \\

\end{tabular}
\vspace{-5ex}
\caption{Left (resp. Middle): scatter plots showing final angle $\phi(\epsilon)$ and opening $\theta_1(\epsilon)$ for $\rho =-0.7$ and $\SCALzero$ (resp. $\SCALtwo$). Every color is for a fixed $\delta$ and variable $\epsilon$. Right: Scatter plot showing $(\phi(\SCALzero), \phi(\aTcheb))$.}
\label{fig:scatterplot}
\end{center}
\vspace{-6ex}
\end{figure*}

Let us focus on the influence of the opening angle~$\theta_1$ when the direction angle $\delta$ is smaller than $\pi/4$ (similar results hold for $\delta>\pi/4$ and $\theta_2$).
Fig.~\ref{fig:scatterplot} shows the scatter plots of the final angle $\phi$ as a function of the opening angle $\theta_1$ for different direction angles $\delta \in [0,\pi/4]$. A scatter plot gives a set of values $(\theta_1(\epsilon), \phi(\epsilon))$ for the $\epsilon$-values under study. From Proposition~\ref{thm:angles}, for a given direction angle $\delta$, the opening angle $\thetasub{1}$ belongs to the interval $[\delta - \pi/2, 0]$ for $\SCALzero$, and to the interval $[-\pi/4,0]$ for $\aTcheb$. Independently of the scalarizing function, when the direction angle is between $0$ and around $3 \pi / 16$ (blue color), the value of $\phi$ is highly correlated with the opening angle $\theta_1$.
For such directions, a simple linear regression confirms this observation and allows us to explain the relation between the opening angle and the final angle by means of the following approximate equation: $\phi \approx (c + \pi /4) + c \cdot \theta_1$,
such that $c$ equals $0.05$, $0.2$, and $0.4$ for $\rho=-0.7$, $0$, and $0.7$ respectively.
We emphasize that this is independent of the definition of the scalarizing function, and depends mainly on the property of the lines of equal function values. The previous equation tells us that the lines of equal fitness values are guiding the search process following the gradient direction given by the opening angle in the objective space. 
Fig. \ref{fig:scatterplot} (Right) shows that the obtained final angles are equivalent when the opening angle is the same, even for different direction angles and/or scalarizing functions. In fact, we observe that the final angles obtained are very similar for the scalarizing functions $\SCALzero$ and $\aTcheb$ if $\delta$ is the same for both functions and the $\epsilon$-values are chosen in order to have matching opening angles.
Whatever the $\delta$- and $\epsilon$-values, the points are close to the line $y=x$, which shows that independently of the scalarizing function, the final angle is strongly correlated to the opening angle, and not to a particular scalarizing function. Also, the opening of the lines of equal function values have more impact on the dynamics of the search process than the direction angle alone. In this respect, the opening angle should be considered as a key feature to describe and understand the behavior of scalarizing search algorithms.

\vspace{-2ex}
\section{Global Search Behavior}\vspace{-2ex}

\begin{figure}[t]
\begin{center}
\begin{boxedminipage}{0.5\textwidth}
\includegraphics[width=0.5\textwidth]{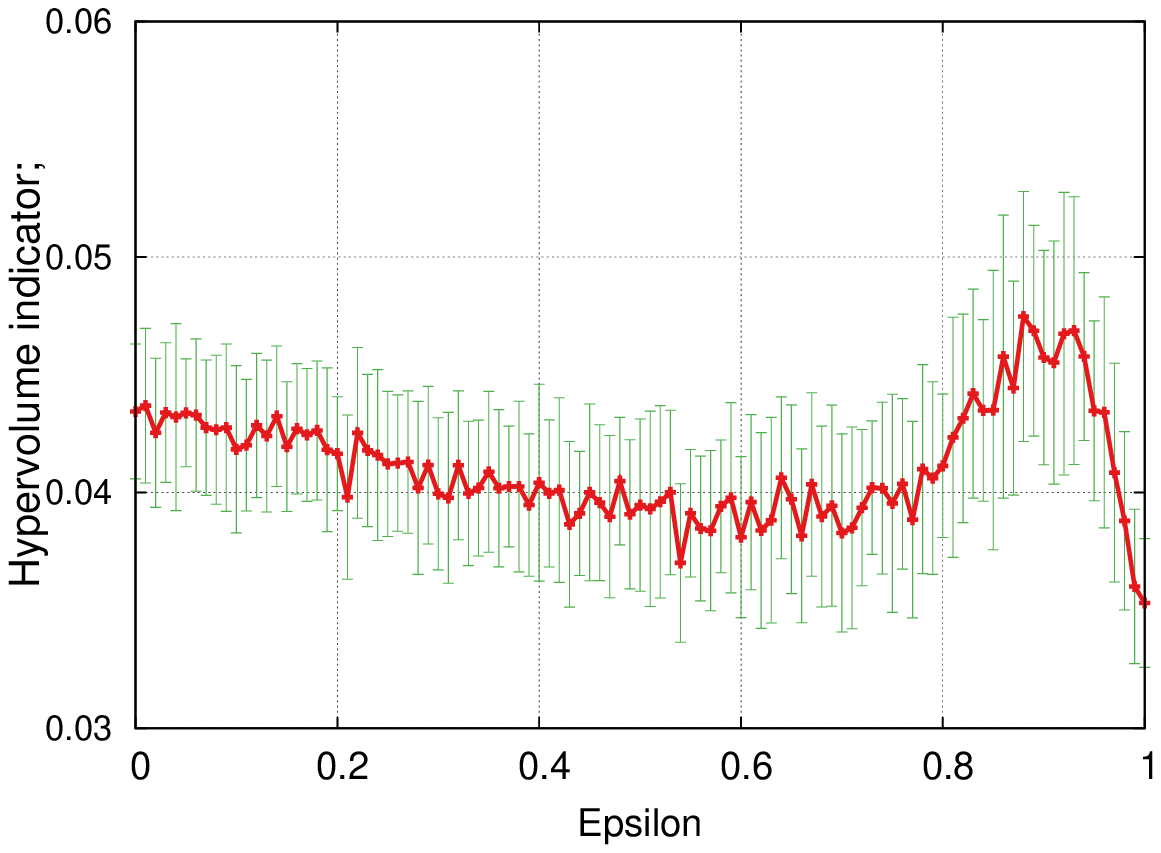}
\hspace{-1.1ex}
\includegraphics[width=0.5\textwidth]{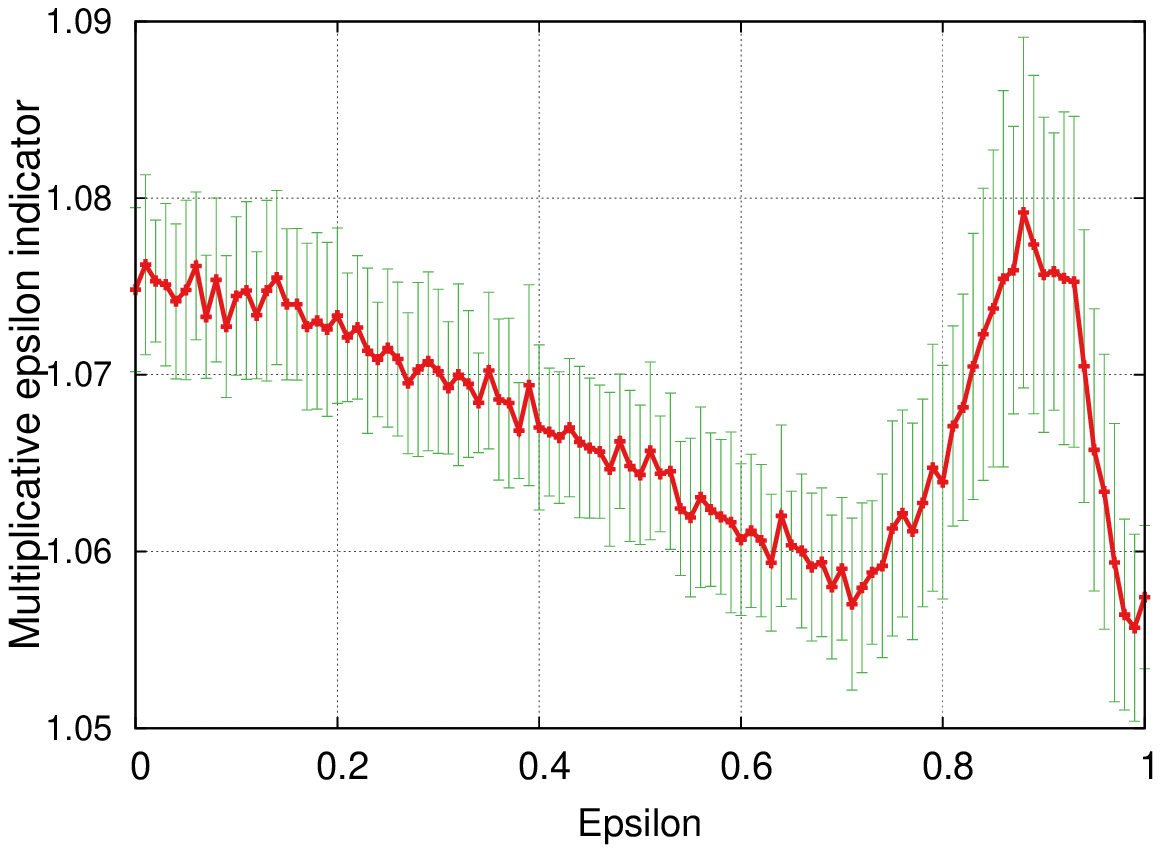}
\end{boxedminipage}
\hspace{-1.1ex}
\begin{boxedminipage}{0.5\textwidth}
\includegraphics[width=0.5\textwidth]{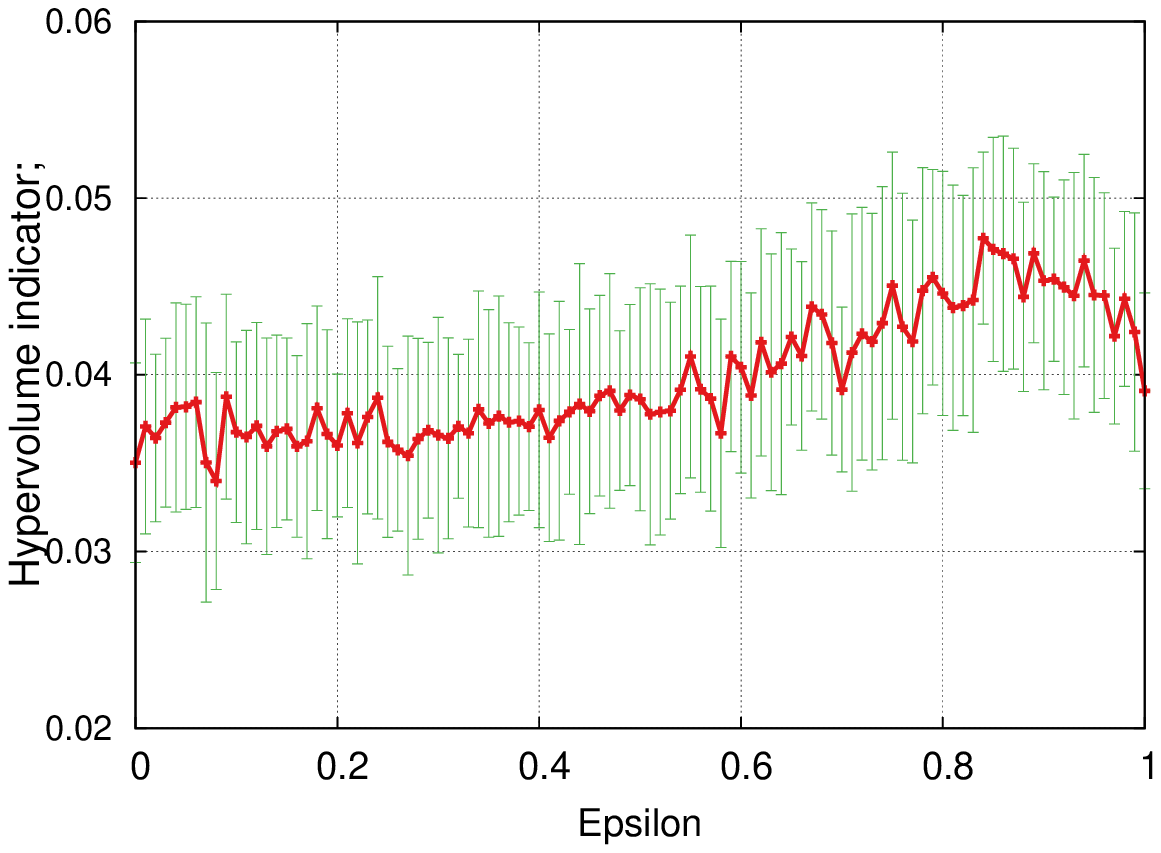}
\hspace{-1.1ex}
\includegraphics[width=0.5\textwidth]{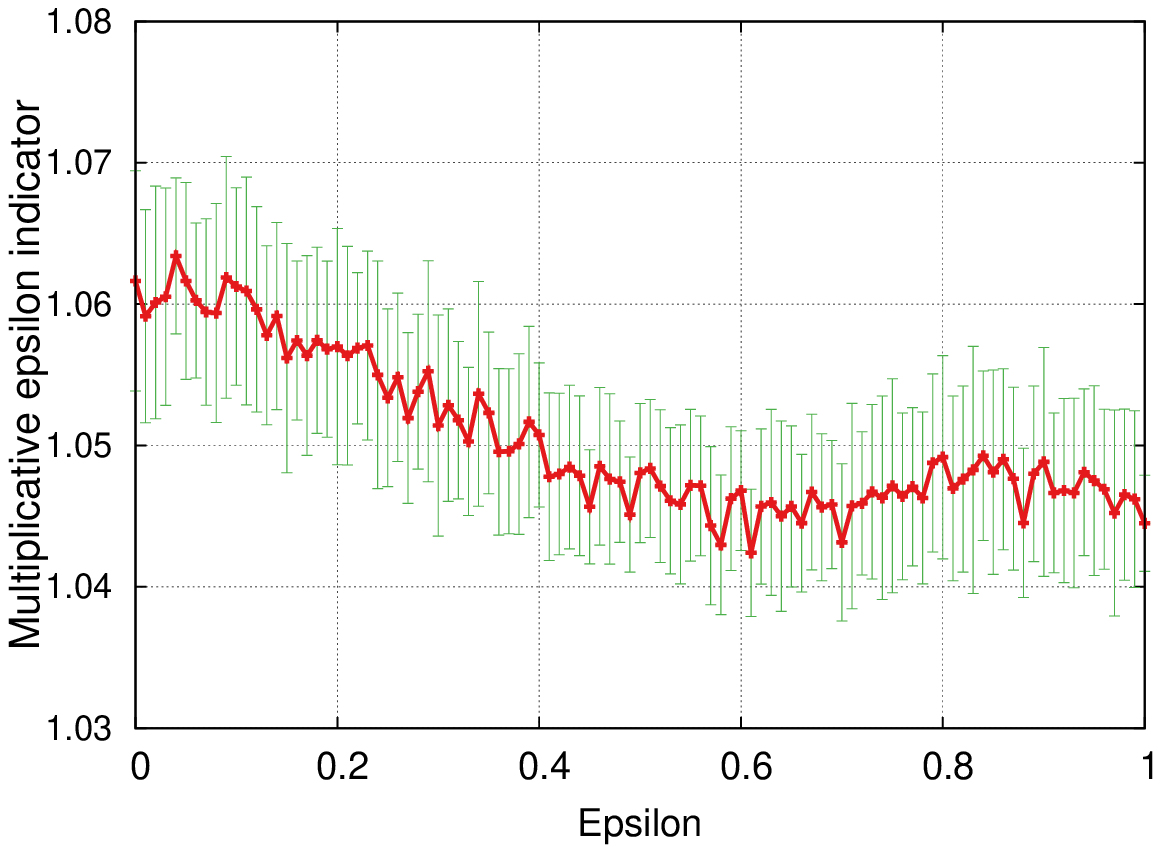}
\end{boxedminipage}

\vspace{-2ex}\caption{Column 1 and 3 (resp. 2 and 4) depict the hypervolume (resp. epsilon) indicators for scalar function \SCALzero. Left (resp. right): objective correlations $\rho=-0.7$ (resp. $\rho=0.7$). 
\vspace{-6ex}}
\label{fig:indicators}
\end{center}
\end{figure}

In the previous section, we considered every single $(1+\lambda)$-EA separately. However, the goal of a general-purpose decomposition-based algorithm is to compute a set of solutions approximating the whole Pareto front. In this section, we study the quality of the set obtained when combining the solutions computed by different configurations of the scalarizing functions. A natural way to do so is to use the same $\epsilon$-value for all direction angles. Fig.~\ref{fig:indicators} illustrates the relative performance, in terms of hypervolume difference  and multiplicative epsilon indicators~\cite{ztlf2003a}, when considering such a setting and aggregating the solutions from the different weight vectors.

The hypervolume reference point is set to the origin, and the reference set is the best-known approximation for the instance under consideration.

Over all the considered \rmnk s, we found that the $\epsilon$-values minimizing both indicator-values correspond to those that allow to well distribute the final angles among direction angles (cf. Fig. \ref{fig:angledeviation}) independently of the considered scalarizing function. Some differences can however be observed depending on the considered indicator, especially for the most correlated instances as illustrated in Fig.~\ref{fig:indicators}. To explain the difference of optimal $\epsilon$-values between both indicators, we remark that the lexicographically optimal regions of the Pareto front approximation have a higher impact on the hypervolume indicator value, due to the setting of the reference point.  For instance, for $\rho=0.7$, the smallest $\epsilon$-values concentrate the final angles to the extreme of the Pareto front, which allows to obtain better results in terms of hypervolume. Contrarily, the epsilon indicator values are better when the final angles are well-distributed around $\pi/4$.

Moreover, \WS is found to be in general competitive with respect to other fixed $\epsilon$-values. This observation might suggest that \WS is the best-performing parameter setting, since every different direction angle leads to a different final angle. Nevertheless, the diversity of final angles is not the only criterion that can explain quality. The efficiency of the $(1+\lambda)$-EA with respect to the single-objective problem implied by the scalarizing function is also crucial. In Fig.~\ref{fig:angledeviation}, we observe that the $\epsilon$-value exhibiting the minimal average deviation to best is not necessarily the same for every direction. We also observe that for direction angles in the middle of the weight space, the final angles obtained for different $\epsilon$-values can end up being very similar. Thus, it might be possible that, by choosing different $\epsilon$-values for different directions, one can find a configuration for which final solutions are diverse, but also closest to the Pareto front.
Indeed, we can observe a significant difference between the non-uniform case where the scalarizing function \SCALzero (or \SCALtwo) is configured with an $\epsilon$ providing the best deviation to best for every direction, and the situation where $\epsilon$ is the same for all directions. As shown in Table~\ref{tab:nonuniformperf}, such non-uniform configurations are both substantially better than \Tcheb and also competitive compared to \WS. 
We only show the performance of the above non-uniform configuration in order to illustrate how choosing different $\epsilon$-values can improve the quality of the resulting approximation set. However, this particular non-uniform configuration might not be `optimal'. In other words, finding the `best' parameter configuration in a setting where $\mu$ independent single $(1+\lambda)$-EAs are considered, can itself be formulated as an optimization problem with variables $\epsilon$ and $\delta$; such that direction angles in the optimal configuration might not necessarily be pairwisely different.

\begin{table}[t]
\caption{Comparison of \WS, \Tcheb, and non-uniform \SCALzerostar and \SCALtwostar configured with $\epsilon$-values giving the best deviation w.r.t every direction. The number in braces shows the number of other algorithms that statistically outperform the algorithm under consideration w.r.t. a given indicator and a Mann-Whitney signed-rank statistical test with a $p$-value of $0.05$ (the lower, the better).\vspace{-2ex}}
\label{tab:nonuniformperf}
\begin{scriptsize}
\begin{center}
\begin{tabular*}{\textwidth}{@{\extracolsep{\fill}}c||cccc|cccc}
 \multicolumn{1}{c}{} & \multicolumn{4}{c}{Avg. hypervolume difference ($\times 10^{-1}$)} & \multicolumn{4}{c}{Avg. multiplicative epsilon} \\\hline\hline
$\rho$ & \WS & \Tcheb & \SCALzerostar & \SCALtwostar & \WS & \Tcheb & \SCALzerostar & \SCALtwostar \\ \hline
$-0.7$ & 0.353 (2) & 0.434 (3) & 0.324 (\textbf{0}) & 0.307 (\textbf{0}) & 1.057 (\textbf{0}) & 1.075 (3) &1.059 (\textbf{0}) & 1.057 (\textbf{0}) \\
$0.0$ & 0.418 (2) & 0.458 (3) & 0.357 (1) & 0.322 (\textbf{0}) & 1.056 (\textbf{0}) & 1.084 (3) & 1.062 (1) & 1.064 (1) \\
$0.7$ & 0.391 (3) & 0.350 (2) & 0.303 (\textbf{0}) & 0.292 (\textbf{0}) & 1.044 (\textbf{0}) & 1.062 (3) & 1.047 (1) & 1.047 (1) \\
\end{tabular*}
\end{center}
\end{scriptsize}
\vspace{-6ex}
\end{table}

\vspace{-1.5ex}
\section{Open(ing) (Re)search Lines}
\label{sec:openissues}
We presented an extensive empirical study that sheds more light on the impact of scalarizing functions within decomposition-based evolutionary multiobjective optimization. Our results showed that, given a weighting coefficient vector and a relative importance of the weighted sum and the Chebychev term in the function, it is fundamentally the opening of the lines of equal function values that explicitly guides the search towards a specific region of the objective space.
When combining multiple scalarizing search processes to compute a whole approximation set, these lines play a crucial role to achieve diversity. 
While our results are with respect to a rather simple setting where multiple scalarizing search procedures are run independently, they make a fundamental step towards strengthening the understanding of the properties and dynamics of more complex algorithmic settings. It is our hope that the lessons learnt from our study can highly serve to better tackle the challenges of decomposition-based approaches. They also rise new interesting issues that were hidden by the complex design of well-established algorithms. In the following, we identify a non-exhaustive number of promising research directions that relate directly to our findings. 

\noindent%
\textbf{\ding{202} Improving existing algorithms.} Eliciting the best configuration to tackle a multiobjective optimization problem by decomposition can highly improve search performance. As we demonstrated, similar regions can be achieved using different parameter settings, and the performance could be enhanced by adopting non-uniform configurations. One research direction would be to investigate how such \emph{non-uniform} configurations perform when plugged into existing approaches. 
To our best knowledge, there exists no attempt in this direction, and previous investigations did only consider uniform parameters, which do not necessarily guarantee to reach an optimal performance.

\noindent%
\textbf{\ding{203} Tuning the opening angles.} Generally speaking, the parameters of existing scalarizing functions can simply be viewed as one specific tool to set up the openings of the lines of equal function values. In this respect, other types of opening angles can be considered without necessarily using a particular scalarizing function. This would offer more flexibility when tuning decomposition-based algorithms, e.g., defining the opening angles without being bound to a fixed closed-form definition, but adaptively, with respect to the current search state. We believe that classical paradigms for on-line and off-line parameter setting are worth to be investigated to tackle this challenging issue.

\noindent%
\textbf{\ding{204} Variation operators and problem-specific issues.} In our study, we consider the independent bit-flip mutation operator and bi-objective \rmnk s. In future work, other problem types and search components should be investigated at the aim of gaining in generality---also towards problems with more than two objectives. 

\noindent%
\textbf{\ding{205} Theoretical modeling.} A challenging issue is to provide a framework, abstracting from problem-specific issues, and allowing us to reason about de\-composition-based approaches in a purely theoretical manner. This would enable us to better harness scalarizing approaches and to derive new methodological tools in order to improve our practice of decomposition-based evolutionary multiobjective optimization approaches.

\vspace{-3ex}
\bibliographystyle{splncs}

\end{document}